\documentclass[conference]{IEEEtran}

\makeatletter

\usepackage{blindtext}
\usepackage{eso-pic}

\usepackage{amsmath,amssymb,amsfonts}

\usepackage{graphicx}
\usepackage{pgf-pie,etoolbox}
\usepackage{textcomp}
\usepackage{xcolor}
\def\BibTeX{{\rm B\kern-.05em{\sc i\kern-.025em b}\kern-.08em
    T\kern-.1667em\lower.7ex\hbox{E}\kern-.125emX}}
    \usepackage{float}
\usepackage{float}
\usepackage{algpseudocode}
\usepackage{placeins}
\usepackage{enumitem}
\usepackage{multicol}

\usepackage{algorithm}
\usepackage{tikz}
\usetikzlibrary{shapes}
\patchcmd\pgfpie@slice
  {node {\scalefont{#3}\beforenumber#3\afternumber}}
  {node[/every only number node/.try] {\scalefont{#3}\beforenumber#3\afternumber}}{}{}

\usepackage{caption}

\usepackage{titlesec}

\usepackage{graphicx}
\usepackage{amsmath}
\renewcommand{\arraystretch}{2.1}
\usepackage{svg}

\usepackage{setspace}
\usepackage{eso-pic}

\begin{document}

\title{Novel Deep Learning Pipeline for Automatic Weapon Detection}
\author{\IEEEauthorblockN{1\textsuperscript{st} Haribharathi.S}
\IEEEauthorblockA{\textit{Department of Computing Technologies,} \\
\textit{Faculty of Engineering and Technology,}\\
\textit{SRM Institute of Science and Technology,}\\
Tamil Nadu-603 203, India \\
hs7886@srmist.edu.in
}
\and
\IEEEauthorblockN{1\textsuperscript{st} Vijay Arvind.R}
\IEEEauthorblockA{\textit{Department of Computing Technologies,} \\
\textit{Faculty of Engineering and Technology,}\\
\textit{SRM Institute of Science and Technology,}\\
Tamil Nadu-603 203, India \\
va0149@srmist.edu.in}
\and
\IEEEauthorblockN{2\textsuperscript{nd} Pawan Ragavendhar V}
\IEEEauthorblockA{\textit{Department of Computing Technologies,} \\
\textit{Faculty of Engineering and Technology,}\\
\textit{SRM Institute of Science and Technology,}\\
Tamil Nadu-603 203, India \\
pr4464@srmist.edu.in}
\and
\IEEEauthorblockN{3\textsuperscript{rd} G.Balamurugan}
\IEEEauthorblockA{\textit{Department of Computing Technologies,} \\
\textit{Faculty of Engineering and Technology,}\\
\textit{SRM Institute of Science and Technology,}\\
Tamil Nadu-603 203, India \\
balamurg1@srmist.edu.in}
}

%
%

%
%

%

\maketitle
\begin{abstract}
Weapon and gun violence have recently become a pressing issue today. The degree of these crimes and activities has risen to the point of being termed as an epidemic. This prevalent misuse of weapons calls for an automatic system that detects weapons in real-time. Real-time surveillance video is captured and recorded in almost all public forums and places. These videos contain abundant raw data which can be extracted and processed into meaningful information. This paper proposes a novel pipeline consisting of an ensemble of convolutional neural networks with distinct architectures. Each neural network is trained with a unique mini-batch with little to no overlap in the training samples. This paper will present several promising results using multiple datasets associated with comparing the proposed architecture and state-of-the-art (SoA) models. The proposed pipeline produced an average increase of 5\% in accuracy, specificity, and recall compared to the SoA systems.
\end{abstract}

\section{Introduction}
Gun violence and knife-aided crimes have significantly increased in the past decade \cite{sturup2019increased}. The leading cause of this new fold surge is mainly due to the legality of owning these weapons. An estimated 8 million new firearms are manufactured daily, and almost 650 million civilians around the globe own firearms for personal use \cite{karp2018estimating}. As per statistics, close to 2000 people get injured, and more than 500 people die every day due to gun violence, and 44\% of all homicides globally involve firearms. Moreover, shootings in schools and other public forums have increased vastly in the past decade, posing threat and danger to young people.\\
On the other hand, knife is legal almost everywhere globally, primarily for cutlery-related activities. But for the past decade, they have been used to injure and hurt people. This disturbing situation calls for innovation to tackle and reduce risk and to ensure a safe life. The current weapon detection (WD) models are inaccurate and slow. Closed Circuit Television (CCTV) cameras are installed almost in all public places and forums for surveillance, but these cameras still need human supervision to monitor activity in real-time. Moreover, this process of monitoring is tedious and time-consuming. Hence, an accurate and reliable innovation is proposed to detect weapons in real time.

Convolutional neural networks (CNN) are widely used for image classification. A CNN extracts feature from the training images and predicts using its trained knowledge \cite{albawi2017understanding}. Although a single CNN can identify and detect objects, it is problematic if it over-fits and makes an error by identifying an object falsely. This would raise a false alarm and question the credibility of the model. Thus this research focuses on implementing a group of N neural networks ($\kappa$), also called an ensemble \cite{dietterich2000ensemble}, and analyzing the credit of the same. The architectures of the base models (BM) are subsequently discussed in the upcoming sections.

 The research is structured as follows, section 2 reviews the previous work in this area and outlines its associated results, Section 3 presents the objectives to be achieved in this work, followed by methodology and the system architecture in Section 4. Section 5 publishes the results obtained in this research, and analysis of the same will be done using suitable comparisons with SoA systems to prove the viability of our research. Section 8 concludes the research and recommends possible work to be carried out as an extension. 

\section{Research Gap}
The major mainstream object detection models employ many deep learning-based models, such as Faster R - CNN, Retina Net, YOLOv7, and others. When considering weapons, one trained model accounts for a significant portion of these models. R. Girshick et al. \cite{girshick2014rich} developed an R-CNN model based on training with real and synthetic images, which indeed increases the accuracy of the model. However, it requires roughly two seconds to detect the target objects in real time. Jinhao Yuan et al. \cite{yuan2022detection} proposed a system that primarily uses the YOLOv7 model for detecting prohibited items and achieves an accuracy of 86.55\%. Numerous tests have found that YOLOv7 is inaccurate compared to other models, such as Faster R-CNN or RetinaNet. Harsh Jain et al. \cite{jain2020weapon} presented a solution to detect weapons based on Single Shot Detector (SSD) and R-CNN. The proposed system took 0.736 s/frame, but only produced an accuracy of 84.6\% and 73.6\%, which is not acceptable for a system that is to be implemented in real-time. To tackle this setback, Yunqi Cui et al. \cite{cui2019automated} proposed a weapon detection system based on RetinaNet as the backbone. Although their model's performance is good and has an acceptable reaction time, the model was trained and tested with generated synthetic images rather than real-world examples, making it questionable when implemented in real time.\\ Hence there exists a significant research gap in the field of WD. This research proposes a reliable and accurate Weapon Detection Pipeline (WDP) for real-time weapon detection. The architecture of the pipeline is discussed in the upcoming sections.
\section{Data \& Preprocessing}\FloatBarrier
Weapon Detection Dataset (WDD) \cite{WDD}, Gun Dataset (GD) \cite{GD}, and the Gun Object Detection dataset (GDD) \cite{GDD} were considered for training and validating the BM for weapon detection. WDD contains images of individuals wielding weapons such as guns and knives. In comparison, the GDD and GD contain images of guns of different types and makes. The datasets are then combined to form a single dataset (A) containing a total of 9391 images. The preprocessing engine ($\beta$) included four sequential stages: Grayscale conversion, Normalization, Scaling, and Train-test split. The images were converted to grayscale to give more information per pixel to learn from for the base learners (BL). Normalization is done to each example in the dataset to speed up the learning rate. The data is then scaled and split into train and test datasets in the ratio of 75:25. This will leave a total of 7043 images for training the BM and 2348 images for validating the performance of the same.  
\begin{equation}
Data (A) \longrightarrow [WDD \hspace{.3cm} GDD \hspace{.3cm} GD]
\end{equation}

\begin{figure}[h!]
    \centering
    \includegraphics[width=0.5\textwidth,height=50mm]{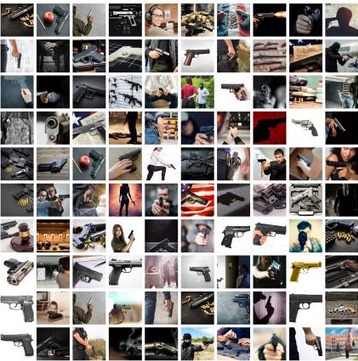}
    \caption{Samples from the dataset}
    \label{fig:my_label}
\end{figure}

The training images are then split into a fixed number of mini-batches ($\alpha_i$) that contain images lesser than the actual dataset for each batch. X mini-batches ($\alpha_x$) are split, x being the number of BL. Each BL is trained on a unique $\alpha_i$ and a randomized batch of samples from all batches. Each $\alpha_i$ is ensured to have a minimum number of examples of all types of weapons with little to no overlap of images between themselves. This ensures that each BL has seen and trained over all weapons, to eliminate out-of-sample. \par

\begin{equation}
A \longrightarrow \alpha_1, \alpha_2, \alpha_3,. . . . . .\hspace{.1cm}\alpha_x 
\end{equation}

\begin{table}[h]
\centering
\caption{Datasets used for training neural networks}
\scalebox{0.9}{
\begin{tabular}{|c|c|c|c|c|}
\hline
 Sr.no & \textbf{Dataset} & \textbf{Total Data} & \textbf{Training Data}  & \textbf{Testing Data}\\
\hline
1 & \textbf{WDD} & 5891 & 4418 & 1473 \\
\hline
2 & \textbf{GD} & 2078 & 1559 & 519 \\
\hline
3 & \textbf{GDD} & 1422 & 1066 & 356 \\
\hline

\end{tabular}}

\end{table}

\section{Methodology}
The system proposed (WDP) is an end-to-end pipeline that takes a frame ($C_i$) as its input and detects weapons if present in the frame. The input $C_i$ is first pushed onto $\beta$ for pre-processing and then fed to the  WDP for detection ($\Delta$). The WDP comprises several interdependent base models (BM), each constructed with distinct architectures and trained in diverse environments and configurations. This increases the WDP's ability to adapt to varying conditions and minimize the impact of an individual model's limitations. WDP was developed with the primary objective of mitigating the prevalence of false predictions that are often observed in the current SoA detection systems. By leveraging the collective power of multiple BMs, the WDP aims to enhance detection performance by reducing false positives and negatives.
\begin{equation}
    \Delta = [\alpha_x\hspace{+.2cm}\beta\hspace{+.2cm}\gamma_x]
\end{equation}

\begin{figure}[h!]
    \centering
\includegraphics[width=0.5\textwidth,height=50mm]{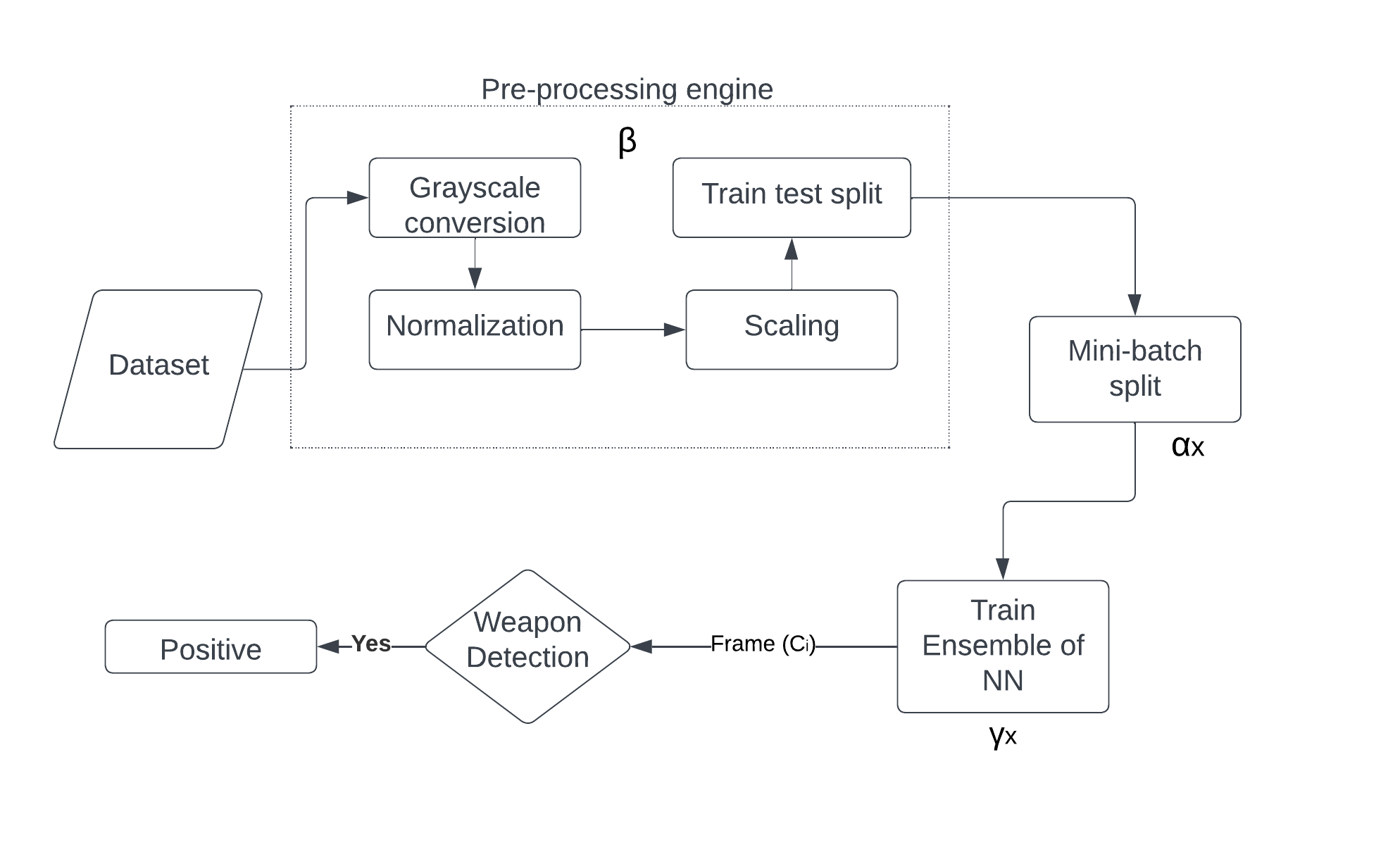}
    \caption{Pipeline Architecture}
    \label{fig:my_label}
\end{figure}

\subsection{Neural network construction}

The construction of $\kappa$ involved designing and implementing a group of interconnected convolutional neural networks capable of working together to improve the accuracy of WDP. Several factors were considered when constructing $\kappa$, like, the architecture of the individual BM used, the optimization algorithm, and the training parameters used in each BM. The architecture of the BM could significantly affect the framework's performance, speed, and accuracy \cite{basha2020impact}. After a series of experiments and simulations, the number of layers for each BM was chosen to be not more than 7, and each BM in $\kappa$ was constructed with varying architectures to make the framework more robust.

The optimization algorithm and training parameters also played a vital role in the construction. Mini batch gradient descent \cite{hinton2012neural} is used as the optimization algorithm in all the BM, which involves updating the weights and biases of the networks based on the error produced on a small batch of training examples. The number of epochs was set to be the same for all BM, one of the most critical training parameters that could affect the model's performance. 
\begin{table}[htbp]
    \centering
    \caption{Comparison of SoA models}
    \label{tab:object_detection}
    \scalebox{0.8}{
    \begin{tabular}{|c|c|c|c|c|c|}
    \hline
        \textbf{Model} & \textbf{Faster R-CNN} & \textbf{RetinaNet} & \textbf{YOLOv7} & \textbf{R-FCN} & \textbf{Ensemble} \\
        \hline
        \textbf{Parameters} & 20M & 32M & 25M & 15M & 92M \\
        \hline
        \textbf{Time (ms)} & 45.6 & 61.4 & 44.2 & 54.5 & 92.3 \\
        \hline
        \textbf{Features} & 1000 & 800 & 1200 & 900 & 1500 \\
        \hline
    \end{tabular}}
\end{table}
\subsection{Ensemble creation}

An ensemble ($\kappa$) was constructed to consolidate predictions from multiple base models (BM) into a single, more accurate detection system. This approach has been demonstrated to be beneficial in enhancing the performance of a machine learning model, particularly for complicated tasks \cite{webb2004multistrategy} such as weapon detection. Each base learner (BL) is trained parallelly with no dependence between each other and in disparate environments. False predictions are significantly mitigated by diverse architectures and minimal similarity between individual $\alpha_i$. The key advantage stems from the observation that when trained in distinct environments, the majority of BL cannot exhibit inaccuracies simultaneously on the same data. Aggregating the outputs of all BMs and considering their mean as the primary output effectively reduced false predictions, rendering the ensemble approach highly suitable for real-time implementation.  \\
 \begin{equation}
Output  = (1/n)\sum_{i=1}^{n}{x_i}
 \end{equation}

\section{Results and Analysis}

\subsection{Metrics}
The metrics used to evaluate the performance of the models are formulated below:

\begin{equation}
Accuracy  = \frac{TP+TN}{TP+TN+FP+FN}
 \end{equation}

 \begin{equation}
Precision  = \frac{TP}{TP+FP}
 \end{equation}

\begin{equation}
Sensitivity  = \frac{TP}{TP+FN}
\end{equation}
\\
$TP$ = Correctly identifies the presence of a weapon
\\
$FP$ =  Incorrectly identifies the presence of a weapon 
\\
$FN$ =  Fails to detect the presence of a weapon
\\
$TN$ = Correctly identifies the absence of a weapon
\\

Measurements like inference time and memory size of the models are also used to evaluate the system in this work. \\
\subsection{Comparison with Base Models}
\renewcommand{\arraystretch}{2.4}
\begin{table}[h]
\centering
\caption{Experimental results of BM}
\label{res2}
\scalebox{0.7}{
\begin{tabular}{|c@{\hspace{10pt}} |c @{\hspace{10pt}}|c@{\hspace{10pt}} |c @{\hspace{10pt}}|c @{\hspace{10pt}}|c@{\hspace{10pt}} |c@{\hspace{10pt}}|}
\hline
\textbf{Metrics} & \textbf{BM1} & \textbf{BM2} & \textbf{BM3} & \textbf{BM4} & \textbf{BM5} & \textbf{$\kappa$}\\
\hline
Accuracy & 0.8902 & 0.8742 & 0.8624 & 0.8452 & 0.8234 & 0.9463\\
\hline
Precision & 0.8988 & 0.8765 & 0.8542 & 0.8411 & 0.8251 &0.9578\\
\hline
Sensitivity & 0.8762 & 0.8432 & 0.8654 & 0.8341 & 0.8227 &\textbf{0.9692}\\
\hline
\end{tabular}
}

\end{table} 

To test the superiority in performance of the $\kappa$ compared to its base model's performance, we evaluated $\kappa$ and all of its constituent BM in separate environments. The BM's was first evaluated individually with no dependence and relation to other networks. Then, the performance of $\kappa$ (after establishing dependency between the base models) was evaluated. In this experiment, five base models (BM1, BM2, BM3, BM4, BM5) were used in the ensemble to achieve a reliable and high-performance pipeline with minimal computation. If the number of BM was chosen lesser than 5, then the accuracy does not supersede the SoA models. If more than 5 BM were chosen, a higher accuracy could be observed, but the increased computation time makes the pipeline unsuitable for real-time application. Each BM was built with an equal number of hidden layers; in this research, seven hidden layers were used, and three evaluation metrics were used in this experiment: accuracy, precision, and sensitivity. Sensitivity was primarily chosen to assess the ability of the models to predict true positives correctly and to reduce false positives. The results of the experiments are shown in Table~\ref{res2}.
\FloatBarrier
\begin{figure}
\includegraphics[width=0.5\textwidth,height=50mm]{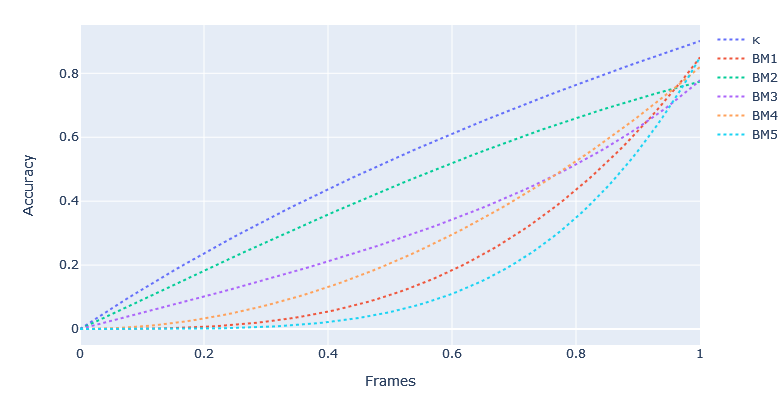}
    \caption{Uniformly distributed performance depiction of BM}
 \end{figure}
According to the results in Table~\ref{res2}, the pipeline ($\kappa$) exhibited an accuracy of 94.63\%, precision of 95.78\%, recall of 96.78\%, and sensitivity of 96.92\%. On the other hand, results depicted that the performance of the other neural networks varied due to their disparate architectures but evidently trailed behind in terms of performance from $\kappa$. BM1 had the second-highest accuracy at 89.02\%, followed by BM2 and BM3 at 87.42\% and 86.24\%, respectively. Furthermore, BM4 and BM5 both got accuracy ratings of 84.52\% and 82.34\%. Hence, the pipeline integrated with $\kappa$ as its backbone is more credible and suitable for real-time application due to its high accuracy and comparatively low inference time. 

\begin{figure*}

\includegraphics[width=\textwidth,height=60mm]{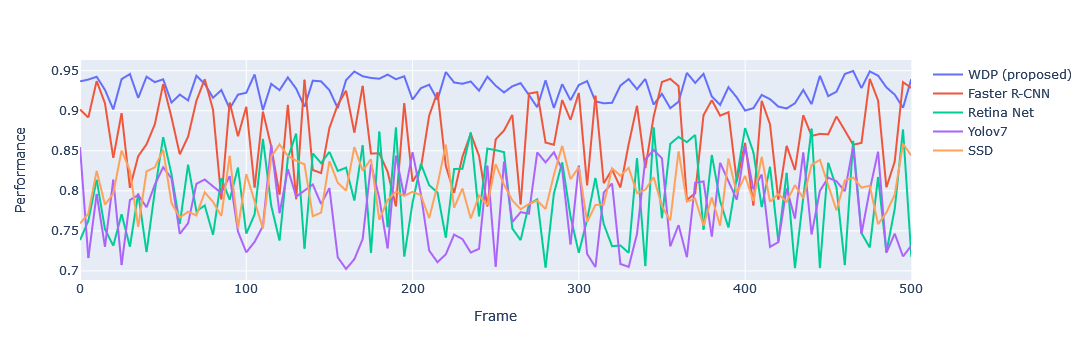}
\centering
\caption{Performance comparison of SoA models}\label{b101}
    
\end{figure*}
\FloatBarrier
\subsection{Result analysis with SoA Models}

The above graph (Fig.~\ref{b101}) was plotted to demonstrate the efficiency of the $\kappa$ pipeline compared to the existing SoA models with performance in the y-axis and the number of test samples in the x-axis. Five hundred test samples that were diverse and unseen by the models were used to evaluate their performance. The performance of four models, Faster R-CNN(orange) \cite{girshick2014rich}, Retina net(cyan) \cite{cui2019automated}, YOLOv7(purple) \cite{yuan2022detection}, WDP(blue), SSD(yellow) \cite{jain2020weapon} are plotted. It can be seen that the ensemble based $WDP$ framework outperformed all the other SoA techniques by a clear margin. RetinaNet and SSD emerged as the least performing models among the five compared models. Both Faster R-CNN and YOLOv7 displayed promising results and were the closest to our proposed pipeline in terms of performance, but was not close enough to outperform $\kappa$ at any point of the experiment. In terms of stability and accuracy, the $\kappa$ consistently outperformed the other models, making it highly suitable for real-time applications. Despite having a significantly larger number of trained parameters (92M) compared to other SoA models, the inference time (0.87) of the pipeline using $\kappa$ shows minimal degradation, with only YOLOv7 performing slightly faster (0.47). Its superior performance and reliability validate its potential for practical implementation.

\begin{table}
\centering
  \caption{Experimental results}
\scalebox{0.9}{
\begin{tabular}{|c|c|c|c|c|c|}
\hline
\textbf{Models} & \textbf{Accuracy} & \textbf{Precision}  & \textbf{Sensitivity} & \textbf{Time (sec)}  \\
\hline
Faster R-CNN \cite{girshick2014rich}& 0.856 & 0.847  & 0.869 & 1.74\\
\hline
Retina net \cite{cui2019automated} & 0.786 & 0.825 & 0.790 & 0.98\\
\hline
YOLOv7 \cite{yuan2022detection} & 0.927 & 0.918  & 0.912 & 0.46 \\
\hline
SSD \cite{jain2020weapon} & 0.798& 0.809  & 0.793 & 1.1\\
\hline
WDP & 0.942 & 0.957  & \textbf{0.969} & 0.87 \\
\hline
\end{tabular}}
\end{table}

\subsection{Pipeline Analysis}

Although the results and the computational resources consumed by $\kappa$ are analyzed and reported throughout the work, it is equally important to look into the overall consumption of inference time of the whole pipeline. As inferred from the figure below, the pipeline inference time is primarily occupied by the $\kappa$ framework. This is due to the complexity of the detection and due to its real-time application. It is worth noting that the framework $\kappa$ outperformed all the SoA models in terms of inference time, as stated in the previous section. Normalization and scaling in the pre-processing engine ($\beta$) take up the remaining time. The inference times were derived from the previous experiment with other SoA models.

\renewcommand{\arraystretch}{0.8}
\begin{tikzpicture}[scale=0.8]
  \definecolor{color1}{RGB}{149,149,149}      
  \definecolor{color2}{RGB}{64,64,64} 
  \definecolor{color3}{RGB}{211,211,211} 

  \pie[
    /tikz/every pin/.style={align=center},
    every only number node/.style={text=white},
    text=pin,
    rotate=240,
    explode=0,
    color={color1,color2,color3}
  ]{
    28.6/Normalization,
    25.4/Scaling,
    46/Detection
  }

\end{tikzpicture}

\section{Conclusion}
This research presented a novel framework for detecting weapons in real-time using an ensemble of convolutional neural networks. From the results obtained in experiment 1, it can be observed that the proposed framework was shown to be effective in increasing the accuracy and reducing false positives and negatives compared to using a single neural network. Experiment 2 shows that the proposed framework surpasses all the other state of the art models by a fair margin and proved to detect weapons faster and more accurately. It is also worth noting that the advancement of weapon detection systems would be highly beneficial to prevent misuse and monitoring the activity of weapon carrying individuals. Further research can be carried out on optimizing the framework to accommodate more neural networks without compromising much on computational resources.  \par

\FloatBarrier
\bibliographystyle{ieeetr}
\bibliography{References}

\begin{thebibliography}{10}

\bibitem{sturup2019increased}
J.~Sturup, A.~Rostami, H.~Mondani, M.~Gerell, J.~Sarnecki, and C.~Edling,
  ``Increased gun violence among young males in sweden: a descriptive national
  survey and international comparison,'' {\em European Journal on Criminal
  Policy and Research}, vol.~25, no.~4, pp.~365--378, 2019.

\bibitem{karp2018estimating}
A.~Karp, ``Estimating global civilian-held firearms numbers,'' 2018.

\bibitem{albawi2017understanding}
S.~Albawi, T.~A. Mohammed, and S.~Al-Zawi, ``Understanding of a convolutional
  neural network,'' in {\em 2017 international conference on engineering and
  technology (ICET)}, pp.~1--6, Ieee, 2017.

\bibitem{dietterich2000ensemble}
T.~G. Dietterich, ``Ensemble methods in machine learning,'' pp.~1--15, 2000.

\bibitem{girshick2014rich}
R.~Girshick, J.~Donahue, T.~Darrell, and J.~Malik, ``Rich feature hierarchies
  for accurate object detection and semantic segmentation,'' in {\em
  Proceedings of the IEEE conference on computer vision and pattern
  recognition}, pp.~580--587, 2014.

\bibitem{yuan2022detection}
J.~Yuan, N.~Zhang, Y.~Xie, and X.~Gao, ``Detection of prohibited items based
  upon x-ray images and improved yolov7,'' vol.~2390, no.~1, p.~012114, 2022.

\bibitem{jain2020weapon}
H.~Jain, A.~Vikram, A.~Kashyap, A.~Jain, {\em et~al.}, ``Weapon detection using
  artificial intelligence and deep learning for security applications,''
  pp.~193--198, 2020.

\bibitem{cui2019automated}
Y.~Cui and B.~Oztan, ``Automated firearms detection in cargo x-ray images using
  retinanet,'' in {\em Anomaly detection and imaging with X-Rays (ADIX) IV},
  vol.~10999, pp.~105--115, SPIE, 2019.

\bibitem{WDD}
AnkanSharma, ``Weapon detection dataset,'' 2021.

\bibitem{GD}
Serkan, ``Gun dataset,'' 2021.

\bibitem{GDD}
S.~Sasank, ``Gun object detection,'' 2019.

\bibitem{basha2020impact}
S.~S. Basha, S.~R. Dubey, V.~Pulabaigari, and S.~Mukherjee, ``Impact of fully
  connected layers on performance of convolutional neural networks for image
  classification,'' {\em Neurocomputing}, vol.~378, pp.~112--119, 2020.

\bibitem{hinton2012neural}
G.~Hinton, N.~Srivastava, and K.~Swersky, ``Neural networks for machine
  learning lecture 6a overview of mini-batch gradient descent,'' {\em Cited
  on}, vol.~14, no.~8, p.~2, 2012.

\bibitem{webb2004multistrategy}
G.~I. Webb and Z.~Zheng, ``Multistrategy ensemble learning: Reducing error by
  combining ensemble learning techniques,'' {\em IEEE Transactions on Knowledge
  and Data Engineering}, vol.~16, no.~8, pp.~980--991, 2004.

\end{thebibliography}

\end{document}